# ExpDNN: Explainable Deep Neural Network

Chi-Hua Chen

In recent years, deep neural networks have been applied to obtain high performance of prediction, classification, and pattern recognition. However, the weights in these deep neural networks are difficult to be explained. Although a linear regression method can provide explainable results, the method is not suitable in the case of input interaction. Therefore, an explainable deep neural network (ExpDNN) with explainable layers is proposed to obtain explainable results in the case of input interaction. Three cases were given to evaluate the proposed ExpDNN, and the results showed that the absolute value of weight in an explainable layer can be used to explain the weight of corresponding input for feature extraction.

*Introduction*: Several deep neural networks with fully-connected layers were developed for various applications (e.g. prediction, classification, pattern recognition, etc.)[1]. However, the weights in these deep neural networks are difficult to be explained, and the important inputs cannot be extracted. Therefore, this study proposes an explainable deep neural network (ExpDNN) which includes explainable layers for extracting key inputs and features. The explainable layer is a special hidden layer without biases in the ExpDNN, and a one-by-one connection between each two neurons in the input layer and the explainable layer of ExpDNN, respectively. For instance, only one connection with an explainable weight ($w_i$) is built between the $i$-th neuron in the input layer and the neuron in the $i$-th explainable layer, so there are $n$ connections between the input layer and the explainable layer in the ExpDNN which includes $n$ inputs. The fully-connected layers could be considered for other multiple layers to model interaction effects. After training the ExpDNN, the value of $w_i$ is used to explain the weight of the $i$-th input. The features of inputs can be extracted in accordance with these explainable weights, and some inputs with lower weights can be filtered out for reducing input dimensionality.

*Method*: The proposed ExpDNN is constructed by an input layer with $n$ inputs, $n$ explainable layers, a merge layer, $l$ hidden layers and an output layer with $m$ outputs (shown in Fig. 1). The principal designs of ExpDNN include seven key ideas: (1) a one-by-one connection is built between each two neurons in the input layer and the explainable layer, and the initial weight of connection is one; (2) the activation function in the explainable layer is a linear function; (3) no bias is used in the explainable layer; (4) the merge layer merges these $n$ explainable layers and connects to a fully-connected layer (i.e. the first hidden layer) , and the initial weights of connections between the merge layer and the first hidden layer are set to be one; (5) the activation function in the first hidden layer is a linear function, and the initial values of biases in the first hidden layer are set to be one; (6) the proposed ExpDNN contains more than one hidden layer, and nonlinear functions could be adopted as activation functions in the hidden layers; (7) the absolute values of weights in the explainable layers could be used to explain the important levels of inputs under the condition that a low loss value is obtained by the trained ExpDNN.

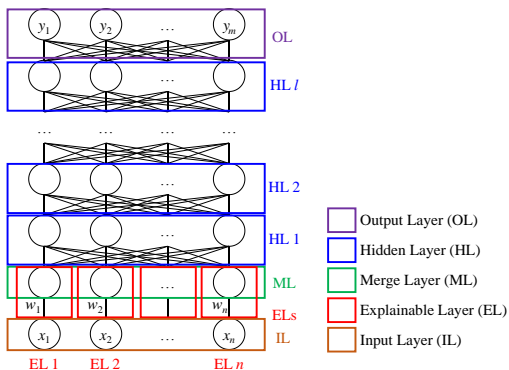

**Fig. 1.** *The structure of ExpDNN*

In the proposed ExpDNN, the neuron in the $i$-th explainable layer ($\upsilon_i$) with a linear activation function can be measured by Equation (1). Furthermore, the function $\phi_k(\bullet)$ denotes the activation function in the $k$-th hidden layer, and the number of neurons in the ($k$-1)-th hidden layer is $n_{k-1}$. The weight of the connection between the $i$-th neuron in the ($k$-1)-th hidden layer and the $j$-th neuron in the $k$-th hidden layer is expressed as $w_{i,j}^{(k)}$, and the bias of the $j$-th neuron in the $k$-th hidden layer is expressed as $b_j^{(k)}$. Therefore, the $j$-th neuron in the first hidden layer ($a_j^{(1)}$) with a linear activation function can be measured by Equation (2). The $j$-th neuron in the $k$-th hidden layer ($a_j^{(k)}$) can be measured by Equation (3), and nonlinear functions could be adopted as activation functions in the $k$-th hidden layer.

$$\upsilon_i = w_i \times x_i. \tag{1}$$

$$a_j^{(1)} = \phi_1\left(\sum_{i=1}^{n} w_{i,j}^{(1)} \times \upsilon_i + b_j^{(1)}\right) = \sum_{i=1}^{n} w_{i,j}^{(1)} \times \upsilon_i + b_j^{(1)}. \tag{2}$$

$$a_j^{(k)} = \phi_k\left(\sum_{i=1}^{n_{k-1}} w_{i,j}^{(k)} \times a_i^{(k-1)} + b_j^{(k)}\right). \tag{3}$$

For obtaining estimated outputs, the weight of the connection between the $i$-th neuron in the $l$-th hidden layer and the $j$-th neuron in the output layer is expressed as $w_{i,j}^{(o)}$, and the bias of the $j$-th neuron in the output layer is expressed as $b_j^{(o)}$. Furthermore, the function $\phi_o(\bullet)$ denotes the activation function in the output layer, and the function $\phi_o(\bullet)$ could be designed in accordance with the applications. For instance, the function $\phi_o(\bullet)$ could be a linear function with the loss function of mean squared error; the function $\phi_o(\bullet)$ could be a sigmoid function with the loss function of binary cross-entropy; the function $\phi_o(\bullet)$ could be a softmax function for the loss function of categorical cross-entropy. Therefore, the $j$-th output in the output layer ($y_j$) can be measured by Equation (4). The Nesterov-accelerated adaptive moment estimation (Nadam) [2] is adopted as an optimizer in this study.

$$y_j = \phi_o\left(\sum_{i=1}^{n_l} w_{i,j}^{(o)} \times a_i^{(l)} + b_j^{(o)}\right). \tag{4}$$

*Case 1:* Case 1 presents a simple application, and Table 1 shows the data in Case 1 [3]. In Case 1, the candidate inputs include $g_1$, $g_2$, $g_3$, $g_4$, and $g_5$; the output is $h$. The high correlations exist among in $g_1$, $g_5$, and $h$ in Case 1. Three subcases which were designed to find the important inputs for the representation of the proposed ExpDNN include: Case 1(1) considered to adopt $g_1$ and $g_2$ as inputs; Case 1(2) considered to adopt $g_1$, $g_2$, $g_3$, and $g_4$ as inputs; Case 1(3) considered to adopt $g_1$, $g_5$, $g_3$, and $g_4$ as inputs. The number of epochs is 60,000, and the loss function of mean squared error is adopted as the loss function of ExpDNN in each subcase.

**Table 1:** The data in Case 1.

| $g_1$ | $g_2$ | $g_3$ | $g_4$ | $g_5$ | $h$ |
|---|---|---|---|---|---|
| 0.1 | 0.1 | 0.3 | 0.5 | 0.7 | 0.3 |
| 0.2 | 0.1 | 0.3 | 0.5 | 0.6 | 0.4 |
| 0.3 | 0.1 | 0.3 | 0.5 | 0.5 | 0.5 |
| 0.4 | 0.1 | 0.3 | 0.5 | 0.4 | 0.6 |
| 0.5 | 0.1 | 0.3 | 0.5 | 0.3 | 0.7 |
| 0.6 | 0.1 | 0.3 | 0.5 | 0.2 | 0.8 |
| 0.7 | 0.1 | 0.3 | 0.5 | 0.1 | 0.9 |

In Case 1(1), Table 2 shows the structure of neural network, and the parameters $g_1$ and $g_2$ denoted the inputs of ExpDNN (i.e. $x_1$ and $x_2$). In the trained ExpDNN, the values of $w_1$ and $w_2$ were 1.2318 and 0.5673, respectively. Therefore, the results showed the important level of $g_1$ was higher than the important level of $g_2$. The high correlation between $g_1$ and $h$ could be extracted by the proposed ExpDNN. Furthermore, Table 3 shows the structure of neural network in Case 1(2) and Case 1(3). In Case 1(2), the parameters $g_1$, $g_2$, $g_3$, and $g_4$ denoted the inputs of ExpDNN (i.e. $x_1$, $x_2$, $x_3$, and $x_4$). In the trained ExpDNN, the values of $w_1$, $w_2$, $w_3$, and $w_4$ were 1.3499, 0.0544, 0.0520, and 0.0515, respectively. Therefore, the results showed the important level of $g_1$ was higher than others, and the high correlation between $g_1$ and $h$ could be extracted by the proposed ExpDNN. In Case 1(3), the parameters $g_1$, $g_5$, $g_3$, and $g_4$ denoted the inputs of ExpDNN (i.e. $x_1$, $x_2$, $x_3$, and $x_4$). In the trained ExpDNN, the values of $w_1$, $w_2$, $w_3$, and $w_4$ were 1.0047, 1.3884, -0.6093, and -0.6140, respectively. Therefore, the results showed the important level list of parameters sorted by the absolute values of weights was $g_5$, $g_1$, $g_4$, and $g_3$. Therefore, the high correlation among in $g_1$, $g_5$, and $h$ could be extracted by the proposed ExpDNN.

**Table 2:** The structure of neural network in Case 1(1).

| Case 1(1) | # of neurons | activation function | use bias |
|---|---|---|---|
| Input Layer | 2 | | |
| Explainable Layer | 2 | linear | False |
| Merge Layer | 2 | | |
| Hidden Layer 1 | 2 | linear | True |
| Hidden Layer 2 | 2 | tanh | True |
| Hidden Layer 3 | 2 | tanh | True |
| Output Layer | 1 | linear | True |

**Table 3:** The structure of neural network in Case 1(2) and Case 1(3).

| Case 1(2) and Case 1(3) | # of neurons | activation function | use bias |
|---|---|---|---|
| Input Layer | 4 | | |
| Explainable Layer | 4 | linear | False |
| Merge Layer | 4 | | |
| Hidden Layer 1 | 4 | linear | True |
| Hidden Layer 2 | 4 | tanh | True |
| Hidden Layer 3 | 4 | tanh | True |
| Output Layer | 1 | linear | True |

*Case 2*: Case 2 presents an application (i.e. an exclusive-OR gate) with input interaction, and Table 4 shows the data in Case 2 [3]. In Case 2, the candidate inputs include $q_1$, $q_2$, $q_3$, and $q_4$; the output is $r$. The input interaction exists between in $q_1$ and $q_2$ for estimating the value of $r$ in Case 2. Two subcases which were designed to find the important inputs with input interaction include: Case 2(1) considered to adopt $q_1$ and $q_2$ as inputs; Case 2(2) considered to adopt $q_1$, $q_2$, $q_3$, and $q_4$ as inputs. The number of epochs is 60,000, and the loss function of binary cross-entropy is adopted as the loss function of ExpDNN in each subcase.

**Table 4:** The data in Case 2.

| $q_1$ | $q_2$ | $q_3$ | $q_4$ | $r$ |
|---|---|---|---|---|
| 0 | 0 | 0 | 1 | 0 |
| 0 | 1 | 0 | 1 | 1 |
| 1 | 0 | 0 | 1 | 1 |
| 1 | 1 | 0 | 1 | 0 |

In Case 2(1), Table 5 shows the structure of neural network, and the parameters $q_1$ and $q_2$ denoted the inputs of ExpDNN (i.e. $x_1$ and $x_2$). In the trained ExpDNN, the values of $w_1$ and $w_2$ were 2.0086 and 2.0086, respectively. Therefore, the results showed the important level of $q_1$ was equal to the important level of $q_2$; both $q_1$ and $q_2$ and were important parameters for estimating the value of $r$. In Case 2(2), Table 6 shows the structure of neural network, and the parameters $q_1$, $q_2$, $q_3$, and $q_4$ denoted the inputs of ExpDNN (i.e. $x_1$, $x_2$, $x_3$, and $x_4$). In the trained ExpDNN, the values of $w_1$, $w_2$, $w_3$, and $w_4$ were 1.9830, 1.9830, 1.0000, and 1.7162, respectively. Therefore, the results showed the important levels of $q_1$ and $q_2$ were higher than others, and the parameter $q_1$ and $q_2$ with input interaction for estimating the value of $r$ could be extracted by the proposed ExpDNN.

*Case 3*: Case 3 presents a practical application of Anderson's Iris data set. Table 7 shows the structure of neural network, and the parameters of sepal length, sepal width, petal length, and petal width denoted the inputs of ExpDNN (i.e. $x_1$, $x_2$, $x_3$, and $x_4$), and the classes of setosa, versicolor, and virginica denoted the outputs of ExpDNN (i.e. $y_1$, $y_2$, and $y_3$). The number of epochs is 60,000, and the loss function of categorical cross-entropy is adopted as the loss function of ExpDNN. In the trained ExpDNN, the values of $w_1$, $w_2$, $w_3$, and $w_4$ were 0.8870, 0.8834, 1.8384, and 1.8925, respectively. Therefore, the results showed the important level list of parameters sorted by the values of weights was petal width, petal length, sepal length, and sepal width. The important features of petal width and petal length for classifying the species of Iris could be extracted by the proposed ExpDNN.

**Table 5:** The structure of neural network in Case 2(1).

| Case 2(1) | # of neurons | activation function | use bias |
|---|---|---|---|
| Input Layer | 2 | | |
| Explainable Layer | 2 | linear | False |
| Merge Layer | 2 | | |
| Hidden Layer 1 | 2 | linear | True |
| Hidden Layer 2 | 2 | tanh | True |
| Hidden Layer 3 | 2 | tanh | True |
| Output Layer | 1 | sigmoid | True |

**Table 6:** The structure of neural network in Case 2(2).

| Case 2(2) | # of neurons | activation function | use bias |
|---|---|---|---|
| Input Layer | 4 | | |
| Explainable Layer | 4 | linear | False |
| Merge Layer | 4 | | |
| Hidden Layer 1 | 4 | linear | True |
| Hidden Layer 2 | 4 | tanh | True |
| Hidden Layer 3 | 4 | tanh | True |
| Output Layer | 1 | sigmoid | True |

**Table 7:** The structure of neural network in Case 3.

| Case 3 | # of neurons | activation function | use bias |
|---|---|---|---|
| Input Layer | 4 | | |
| Explainable Layer | 4 | linear | False |
| Merge Layer | 4 | | |
| Hidden Layer 1 | 4 | linear | True |
| Hidden Layer 2 | 4 | tanh | True |
| Hidden Layer 3 | 4 | tanh | True |
| Output Layer | 3 | softmax | True |

*Conclusions and Future Work*: An ExpDNN with explainable layers is proposed to obtain explainable results in the case of input interaction for extracting features. In the future, the proposed ExpDNN could be applied to various applications.

*Data Availability*: All results were obtained using Python, Tensorflow and Keras. The codes are available at https://github.com/ChiHua0826/ExpDNN.

*Acknowledgments*: This work was supported by National Natural Science Foundation of China (grant 61906043) and Fuzhou University (grants 510730/XRC-18075 and 510809/GXRC-19037).

Chi-Hua Chen (*College of Mathematics and Computer Science, Fuzhou University, Fuzhou City, China.*)

E-mail: chihua0826@gmail.com